\documentclass[letterpaper]{article}

\usepackage[T1]{fontenc}

\usepackage{geometry}
\usepackage{setspace}

\usepackage[style = chem-acs]{biblatex}
\usepackage{graphicx}
\usepackage{float}

\usepackage{chemformula}
\usepackage[version = 4]{mhchem}

\usepackage{amsmath}
\usepackage{amssymb}
\usepackage{booktabs}          
\usepackage{makecell}
\usepackage{microtype}
\usepackage{xcolor}            
\usepackage{hyperref}
\usepackage{url}

\newif\ifdraft\drafttrue

\newif\ifreviewflags\reviewflagstrue
\definecolor{flagblk}{HTML}{C0005A}   
\definecolor{flagmaj}{HTML}{A35200}   
\definecolor{flagmin}{HTML}{0B6E55}   
\definecolor{flagred}{HTML}{6A3DB8}   

\newcommand{\base}{\textsc{Base}}
\newcommand{\chemr}{\textsc{Chem-R}}
\newcommand{\chem}{\textsc{Chem-F}}
\newcommand{\chemdfm}{\textsc{Chem-DFM}}
\newcommand{\reason}{\textsc{Reason}}

\usepackage{authblk}

\author[1]{Tran Nhat Khang Nguyen\textsuperscript{\dag}}
\author[2]{Anh Nguyen\textsuperscript{\dag}}
\author[1]{Obed Issakah}
\author[1]{Janghoon Ock\textsuperscript{*}}
\affil[1]{Department of Chemical and Biomolecular Engineering,
  University of Nebraska--Lincoln, Lincoln, NE 68588, USA}
\affil[2]{School of Computing,
  University of Nebraska--Lincoln, Lincoln, NE 68588, USA}

\title{Functional-Group Knowledge Is Structured in Language Models Before It Is Spoken}
\date{\textsuperscript{*}Email: jock2@unl.edu.
  \textsuperscript{\dag}These authors contributed equally.}

\begin{document}

\maketitle

\begin{abstract}
Language models increasingly serve as the reasoning core of agents for chemistry and materials science, where they must interpret molecular structure before planning syntheses or proposing materials. Yet their outputs show only what they say, not what they know. Given a prompt such as ``The primary functional group in 1-Fluorobutane (CCCCF) is'', a small general-purpose model may answer ``the alkyne group'' instead of an alkyl fluoride. The output alone cannot tell whether the model failed to recognize the fluorine or recognized it and failed to say so. We address this by reading the internal representations of five open-source models (a general-purpose base, three chemistry-trained models, and a reasoning-distilled model) on prompts covering 92 molecules and 20 functional groups. Even without chemistry training, the models organize functional groups by chemical family, and they encode chemical differences as consistent directions. The shift from thioether ($\mathrm{R{-}S{-}R'}$) to thiol ($\mathrm{R{-}SH}$), added to ether ($\mathrm{R{-}O{-}R'}$), lands nearest alcohol ($\mathrm{R{-}OH}$). This structure emerges abruptly in the middle layers, where representations shift from tracking the wording of the prompt to tracking the functional group. The base model names the correct functional group in only 58.5\% of prompts, the lowest of the five, yet it encodes functional-group families and preserves chemically meaningful patterns. Its errors are mostly failures to answer rather than wrong answers. Functional-group knowledge is structured in the model before it is spoken. Chemistry fine-tuning acts less by adding this knowledge than by reorganizing it, making functional groups rather than prompt wording dominate the representation. These results suggest that evaluating chemical language models through outputs alone can underestimate the chemical information they encode. Representation analysis can distinguish a lack of knowledge from a failure to express it, which matters when choosing or adapting models for chemical agents.
\end{abstract}

\section*{Keywords}

Language models, Molecular representation, Functional groups, Organic chemistry, Interpretability




\section{Introduction}
\label{sec:intro}
Large language models are increasingly used in chemistry, valued for their reasoning ability and accessibility through natural language \cite{jablonka2024leveraging, ramos2025review}. They now serve as the reasoning core of agents that generate hypotheses, plan experiments, and carry them out, both in physical laboratories \cite{mbran2024augmenting, boiko2023autonomous, song2025multiagent} and in simulation \cite{zou2025agente, jia2024llmatdesign, chaudhari2026modular}. In many of these applications, the reasoning turns on molecular structural characteristics, such as functional groups. Drug candidates are screened by the groups they present \cite{ceretomassagu2015molecular, ansari2024dziner}, toxicity and metabolic liability are attributed to specific substituents \cite{baell2010new, kirchmair2015predicting, ock2026drugagent}, and in materials design, functional groups govern how molecules bind at a surface \cite{ock2026adsorbagent}. All of these tasks depend on a step that is rarely inspected. Before any reasoning can proceed, the model must read text describing a molecule and recover the structure it denotes. A model that mistakes an ester for a ketone can still plan coherently and fail silently.

In pharmaceutical and materials research, such agents must sometimes run locally on open-weight models, for example when proprietary structures and unpublished data cannot be sent to an external API \cite{openllm, sharma2026openweights}. Choosing and adapting such a model requires knowing why it fails, but chemistry benchmarks score only what a model outputs \cite{white2023assessment, guo2023can, mirza2025framework}. An output score combines what the model knows about chemistry with its ability to state that knowledge in the expected form. A low score may therefore mean that the model lacks the relevant chemistry, or that it has the chemistry encoded but fails to express it. The two readings call for different remedies, namely further chemical training in the first case, and answer-format tuning, few-shot prompting, or constrained decoding in the second. We therefore ask what a model encodes about molecular structure, independent of what it says.

Most work on chemical language models leaves this distinction unexamined. Domain adaptation implicitly treats the chemistry as missing and supplies it by training on scientific literature, as in Galactica \cite{taylor2022galactica}, or on a chemist's reasoning workflow, as in Chem-R \cite{wang2026chem}. Property-prediction studies show that language-model encoders can capture chemically relevant information from text, predicting molecular properties from SMILES strings \cite{chithrananda2020chemberta}, adsorption energies \cite{ock2023catalyst} and band gaps \cite{yeh2026text} from textual descriptions. However, these encoders are fine-tuned on labeled data. The accuracy of their predictions judges them, so they do not reveal what the representation contains before fine-tuning or how it is organized. Agentic systems, in turn, judge chemical competence by the outcome of completed tasks. In each case, the model's internal treatment of chemistry remains a black box.

\textcite{lei2025llamas} instead examined the model's internal representations directly. They showed that the hidden states of a Llama model arrange chemical elements in a geometry that mirrors the periodic table, so that element identity and properties can be recovered from activations rather than from generated text. Their work establishes that a language model maps chemical language into an embedding space that preserves chemically meaningful structure \cite{lei2025llamas, alain2016understanding, belinkov2022probing}. An element symbol, however, is close to a single token, so that geometry rests on units the tokenizer already isolates. Organic chemistry is built on a larger unit, the functional group, which spans several tokens and is defined by what it attaches to a carbon backbone. Therefore, we ask whether language models represent functional groups internally, and how that representation relates to the answers they give.

\begin{figure}[!hbt]
  \centering
  \includegraphics[width=0.9\linewidth]{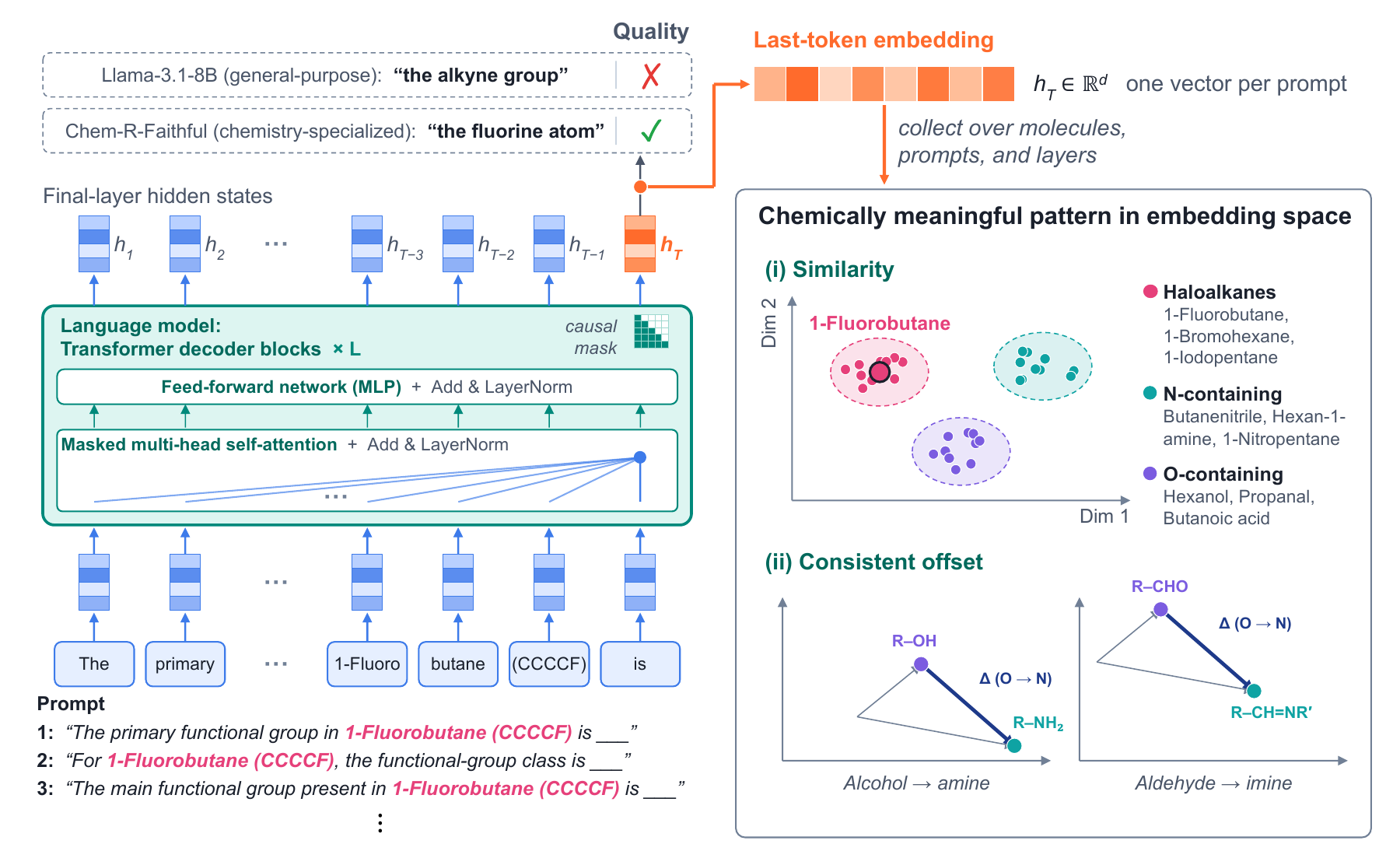}
  \caption{\textbf{Last-token embedding analysis framework.} Language models complete ten declarative prompt templates, each giving a molecule's IUPAC name and SMILES, with a functional-group name (left). We keep the last-token state $h_T$ (orange) at every layer to analyze how the molecule is encoded in the embedding space (right).}
  \label{fig:overview}
\end{figure}

We address this by reading the internal representations of five open-weight 8B language models, a general-purpose base, three chemistry-trained models, and a reasoning-distilled model, on 92 molecules covering 20 functional groups. Each molecule is given by IUPAC name and SMILES string in prompts such as ``The primary functional group in 1-Fluorobutane (CCCCF) is'', which the model completes with a functional-group name (Figure~\ref{fig:overview}). We score these completions and, separately, take the hidden state at the last token of each prompt at every layer. On these embeddings, we measure how functional groups are arranged relative to one another, whether chemical differences behave as consistent directions, and at which layers functional-group information becomes linearly decodable. Comparing the two reveals a gap between what these models encode and what they say.  Even the base model, which is the least accurate at naming functional groups, organizes them by chemical family and encodes chemical differences as consistent directions. This gap also changes how the role of chemistry training should be understood.


\section{Methods}
\label{sec:methods}

\subsection{Molecules and Prompt Templates}

We use 92 small organic molecules covering 20 functional groups, each represented at four backbone lengths from three to six carbons (C3 to C6). The groups fall into five families: the unsubstituted backbone \texttt{none (alkane)}; the halides \texttt{alkyl fluoride}, \texttt{alkyl chloride}, \texttt{alkyl bromide} and \texttt{alkyl iodide}; the oxygen family \texttt{alcohol}, \texttt{ether}, \texttt{aldehyde}, \texttt{ketone}, \texttt{carboxylic acid} and \texttt{ester}; the nitrogen family \texttt{amine}, \texttt{imine}, \texttt{nitrile}, \texttt{amide} and \texttt{nitro}; and the sulfur family \texttt{thiol}, \texttt{thioether}, \texttt{sulfoxide} and \texttt{sulfone}. Seventeen classes have exactly four members, one per chain length, while \texttt{alcohol}, \texttt{amine} and \texttt{thiol} have eight, the 1- and 2-positional isomers at each length. Every molecule is listed with its SMILES string, chain length, and functional group in Section~S1 of the Supporting Information (Table~S1).

Each molecule is instantiated in 10 declarative templates that ask the model to complete a sentence by naming the representative functional group. Each molecule is identified by both its IUPAC name and its SMILES string, and every prompt ends mid-sentence so that the group name is the natural continuation, e.g.\ \texttt{"Propan-1-ol (CCCO) contains the functional group "} and, with the molecule demoted to a fronted prepositional phrase and the target noun changed, \texttt{"For Propan-1-ol (CCCO), the functional-group class is "}. This gives 920 prompts per model. All ten templates are reproduced verbatim in Section~S2 of the Supporting Information (Table~S2).

A single phrasing would make every measurement a property of that phrasing, so the templates vary word order, the syntactic slot the molecule occupies, and the noun used for the target (\emph{group}, \emph{class}, \emph{identity}), while the chemical content stays fixed. Varying the phrasing lets us average out template-specific effects where the analysis calls for it and measure them directly where they matter. Only \texttt{\{iupac\_name\}} and \texttt{\{smiles\}} are interpolated, and each template ends on a trailing space. Prompts are raw completion strings, and no chat template or system message is applied to any model, including the instruction-tuned ones, so the five models read character-identical inputs.

\subsection{Language Models}
We use five open-source language models, all 8B, all with 32 decoder layers over a 4096-dimensional residual stream. \texttt{Llama-3.1-8B} (\base{}) is the base \cite{grattafiori2024llama}. \texttt{Chem-R-8B} (\chemr{}) is chemistry-tuned from it on a chemist's reasoning workflow \cite{wang2026chem}. \texttt{Chem-R-Faithful} (\chem{}) is a reinforcement-learning fine-tune of \chemr{} under a verification-grounded process reward \cite{li2026chemical,wang2026chem}. Its published configuration matches \texttt{Llama-3.1-8B} in vocabulary size, rotary base, width, depth, and context length. \texttt{DeepSeek-R1-Distill-Llama-8B} (\reason{}) is reasoning-distilled from the same base \cite{guo2025deepseek}. \texttt{ChemDFM-v1.5-8B} (\chemdfm{}) is chemistry-tuned on \texttt{Llama-3} rather than \texttt{Llama-3.1} \cite{zhao2025developing, chemdfm15_8b}. The first four share a base, an architecture, and a tokenizer, so every contrast among them varies fine-tuning alone. \chemdfm{} is the exception.

Each of the five models completes all 920 prompts, generating the name of the molecule's functional group. We decode greedily (\texttt{do\_sample=False}, at most 10 new tokens), so that differences between templates are not confounded with sampling noise, and strip the trailing space of each prompt before tokenization. Each continuation is then assigned to exactly one of five outcomes. We first check its format, and a continuation is \emph{malformed} if it lists several options or retracts its own answer. Otherwise, we match it against a lexicon that maps each of the 20 functional-group classes to its accepted names and synonyms, so that, for example, \texttt{-SH} counts as naming a thiol and ``the bromine atom'' as naming an alkyl bromide. A continuation that matches exactly one class is \emph{answered}, and it is either \emph{correct} or \emph{wrong} depending on whether that class is the gold one. A continuation that names only a broader category, such as \texttt{carbonyl} for an ester, is \emph{underspecified}, and one that names no class at all is \emph{non-answer}. Strict accuracy is the share of prompts answered correctly, and lenient accuracy additionally credits superclass-only answers (Table~\ref{tab:paradox}). The detailed rules are provided in the Supporting Information Section~S3.

\subsection{Embedding Extraction}

Throughout this paper, an \emph{embedding} refers to the last-token hidden state. This is the residual-stream vector at the last non-padding token of a prompt, taken at a given layer. The residual stream is the per-token vector that each decoder block reads from and adds to, so the last-token embedding is a single 4096-dimensional vector, distinct from the rows of the model's input embedding table. We use the last token because every prompt template ends just before the functional-group name. At this position, the model has processed the entire prompt and is about to generate its answer, so any information supporting that answer should be present in this state. We record it with forward hooks on all 32 decoder blocks, which gives one embedding per layer for each prompt and lets us track how the representation changes with depth. The embeddings for each layer are stored as a $920 \times 4096$ half-precision tensor, ordered by molecule and then by template, so the first ten rows correspond to the ten templates of the first molecule.

\subsection{Embedding Differences}

Geometric clustermap analysis works with difference vectors rather than raw embeddings. For each molecule, we average its embedding over the 10 templates and subtract the template-averaged embedding of the alkane with the same number of carbons. For example, propanal minus propane is expected to give the aldehyde difference vector at three carbons. Averaging over the ten templates gives one embedding per molecule, and subtracting the alkane removes the backbone the two molecules share. A second correction for the anisotropy that remains is applied at the similarity stage (Section~S7).

Let $x_m \in \mathbb{R}^{D}$, with $D = 4096$, be the template-averaged last-token embedding of molecule $m$, let $a(c)$ be the alkane with $c$ backbone carbons, and let $M_{g,c}$ be the set of molecules with functional group $g$ and carbons $c$. The difference vector for group $g$ at carbon count $c$ is
\begin{equation}
  d_{g,c} = \frac{1}{|M_{g,c}|} \sum_{m \in M_{g,c}} \bigl( x_m - x_{a(c)} \bigr).
  \label{eq:diffvec}
\end{equation}
For \texttt{alcohol}, \texttt{amine} and \texttt{thiol}, $M_{g,c}$ contains both positional isomers, so the average also combines them. This gives 76 difference vectors, one for each of the 19 non-alkane groups at each of the four carbon counts. Matching is on total carbon count rather than on an unbroken chain. In ether, ester, thioether, sulfoxide, and sulfone, the heteroatom interrupts the skeleton, so $d_{g,c}$ for those groups also carries that rearrangement. The alkane itself has no difference vector, since $d_{\text{alkane},c}$ would be zero. It serves as the reference point of this space rather than a point within it.

\subsection{Retrieval from Offset}
This test applies the word-vector analogy question \cite{mikolov2013linguistic} to difference vectors. Given two pairs related by the same chemical change, $a_2 \to a_1$ and $b_2 \to b_1$, we form the query $q = b_2 + (a_1 - a_2)$, rank a pool of candidate difference vectors by cosine similarity to $q$, and record the rank of $b_1$. For example, converting a thioether ($\mathrm{R{-}S{-}R'}$) to a thiol ($\mathrm{R{-}SH}$) involves the same structural change as converting an ether ($\mathrm{R{-}O{-}R'}$) to an alcohol ($\mathrm{R{-}OH}$), so
\begin{equation}
  q = d_{\mathrm{R{-}O{-}R'}} + \bigl( d_{\mathrm{R{-}SH}} - d_{\mathrm{R{-}S{-}R'}} \bigr)
  \label{eq:analogy}
\end{equation}
should land nearest the alcohol vector $d_{\mathrm{R{-}OH}}$. Each quadruple of groups is scored from all four directions, since either pair can supply the offset and either direction can be read. A direction is dropped if it reduces to a plain nearest-neighbour lookup, as when the added and subtracted terms coincide. The three input vectors are excluded from the pool, unless one of them is itself the answer.

We score three axes, each with its own set of quadruples. \emph{Between groups} (20 trials) pairs one substitution against another, such as S$\to$O in thioether/thiol and ether/alcohol, and \emph{down the halide column} (12 trials) varies the halogen among the four alkyl halides. Both use one vector per group, averaged over chain lengths, giving a pool of 16 or 17 candidates after exclusion. \emph{Along the chain-length ladder} (228 trials) varies the backbone from 3 to 6 carbons within a group, with each (group, chain length) vector as a candidate. Both legs of a quadruple must span the same number of steps, since an analogy assumes the two offsets are the same vector. We report hit@1, the share of trials in which the answer ranks first, along with mean and median rank.

\subsection{Probing Test}

At each of the 32 layers, we train a logistic regression probe with $L_2$ regularization to sort embeddings into five families: hydrocarbon, halide, oxygen, nitrogen, and sulfur. To test whether the probe generalizes to functional groups it has never seen, we use leave-one-group-out cross-validation, holding out every molecule of one group at a time and asking the probe to place that group in the right family. The alkane is the only hydrocarbon group, so we never hold it out, since doing so would leave no hydrocarbon examples for training. This gives 19 folds. We standardize the features and weight each class inversely to its size. Within each fold, we choose the regularization strength from $\{10^{-4}, 10^{-3}, 10^{-2}, 10^{-1}\}$ using an inner 3-fold cross-validation that keeps all templates of a molecule in the same split. Performance is measured by balanced accuracy, the average recall across the five families, computed over the pooled out-of-fold predictions.

We compare probe scores against two baselines. The first, \emph{random guess}, assigns one of the five families uniformly at random and gives a balanced accuracy of 0.20. The second, the \emph{character $n$-gram baseline}, is a classifier trained on the prompt strings alone, with no language model involved. It measures how much family information can be recovered from the spelling of a molecule's name and SMILES.

The $n$-gram baseline uses the same 920 unmodified prompts that the models receive. Each prompt is lowercased and split into words at whitespace, and every run of 2 to 5 consecutive characters within a word is counted, so that \texttt{propan-1-ol} contributes \texttt{-ol}, \texttt{1-ol} and \texttt{-1-ol}, among others, while no $n$-gram spans two words. Only $n$-grams occurring in at least two training prompts are kept. Each $n$-gram $t$ in prompt $p$ is weighted by TF-IDF,
\begin{equation}
  w(t, p) = f(t, p) \left[ \ln \frac{1 + N}{1 + n_t} + 1 \right],
  \label{eq:tfidf}
\end{equation}
where $f(t,p)$ is the number of occurrences of $t$ in $p$, $N$ is the number of training prompts, and $n_t$ is the number of training prompts containing $t$. Each prompt vector is then normalized to unit length. This weighting gives the least weight to fragments shared by all prompts, such as template wording, and more weight to distinctive fragments such as \texttt{-ol}. A class-balanced logistic regression with $L_2$ regularization (inverse regularization strength $C = 1$) is then fit on these vectors. Within each fold, the vocabulary, weights and classifier are fit on the training prompts only, and the folds and labels are the same as the probe's.

\section{Results and discussion}
\label{sec:results}

\subsection{Beyond Generation Output}
\label{sec:generation}
Language models differ in how well they describe molecules. Prompted with ``The primary functional group in Hexanamide (CCCCCC(N)=O) is'', \base{} correctly continues ``the amide group''. Given the same template for 1-Fluorobutane (CCCCF), however, it continues ``the alkyne group'', a class the molecule does not contain, whereas \chem{} answers ``the fluorine atom'' (Figure~\ref{fig:overview}). To see how far such differences extend, we first score generation across all molecules and prompt templates.

Generation accuracy varies widely across models, and seemingly chemistry training raises it (Table~\ref{tab:paradox}). Strict accuracy ranges from 0.585 in \base{} to 0.902 in \chemr{}, with \chem{} at 0.843, \reason{} at 0.789 and \chemdfm{} at 0.635, a gain of $+0.259$ for \chem{} over \base{}. These scores come from the greedy continuations of all 920 prompts, scored against the 20 functional-group classes. A prompt counts as \emph{answered} when the continuation names exactly one functional group (Methods Section~\ref{sec:methods}). Strict accuracy is the share of prompts whose answer matches the correct functional group, and lenient accuracy also credits naming only a broader category, such as \texttt{carbonyl} for an ester. Detailed rules are provided in Section~S3.

\providecommand{\stack}[2]{\begin{tabular}[!hbt]{@{}c@{}}#1\\#2\end{tabular}}
\begin{table}[!hbt]
  \centering
  \caption{\textbf{Free generation for functional-group identification.} Accuracy is scored on continuations of up to 10 tokens against 20 functional groups. \emph{Strict acc.}: share of prompts naming exactly the correct functional group. \emph{Template SD}: standard deviation of strict accuracy across templates. \emph{Lenient acc.}: credits broader categories. \emph{Answer rate}: share of prompts naming exactly one group, whether correct or not. \emph{Acc.\ if answered}: strict accuracy among answered prompts. $^{\dagger}$Built on Llama-3 rather than Llama-3.1.}
  \label{tab:paradox}
  \small
  \setlength{\tabcolsep}{3pt}
  \begin{tabular}{lccccc}
    \toprule
    & \multicolumn{5}{c}{Free generation} \\
    \cmidrule(lr){2-6}

    Model & \stack{Strict}{acc.} & \stack{Template}{SD} & \stack{Lenient}{acc.}
      & \stack{Answer}{rate}
      & \stack{Acc.\ if}{answered} \\
      
    \midrule
    Llama-3.1-8B (Base) & 0.585 & 0.172 & 0.667 & 0.637 & 0.918 \\
    Chem-R-8B (Chem-R) & 0.902 & 0.135 & 0.915 & 0.920 & 0.981 \\
    Chem-R-Faithful (Chem-F) & 0.843 & 0.109 & 0.937 & 0.858 & 0.984 \\
    ChemDFM-v1.5-8B$^{\dagger}$ (Chem-DFM) & 0.635 & 0.123 & 0.682 & 0.665 & 0.954 \\
    R1-Distill-Llama-8B (Reason) & 0.789 & 0.100 & 0.847 & 0.828 & 0.953 \\
    \bottomrule
  \end{tabular}
\end{table}

Accuracy also varies with the prompt template. Its standard deviation across templates is 0.172 in \base{} and between 0.100 and 0.135 in the other models. \base{}'s accuracy ranges from 0.196 to 0.880 depending on the template, and its malformed responses are concentrated in one template (template 9; Table~S4). But even after excluding that specific template, \base{}'s accuracy remains at  0.628. \base{} is therefore both the least consistent model across phrasings and less accurate overall. However, how much of its deficit reflects difficulty with the answer format rather than with identifying the functional group cannot be determined from free-generation outcomes alone.

These scores describe outcomes, not the mechanism behind them. An incorrect output alone cannot distinguish between a model that failed to encode the functional group and one that encoded it but failed to express that knowledge in its response. We therefore turn to the representations themselves, analyzing the last-token hidden states to measure what each model encodes, independent of what it says (Figure~\ref{fig:overview}).

\subsection{Chemical Patterns in Embedding Space}
\label{sec:structure}

We compare functional groups through difference vectors, standardized within each model, because raw hidden states are too anisotropic to compare directly (Methods section). In \base{} at layer 31, the mean cosine similarity between arbitrary activations is $0.944 \pm 0.018$ (Table~\ref{tab:geometry}), so unrelated vectors appear similar by default \cite{ethayarajh2019contextual, gao2019representation}. Subtracting the embedding of the alkane with the same carbon count removes the shared backbone and leaves the contribution of the functional group, giving one difference vector per group and chain length (Eq.~\ref{eq:diffvec}).

\begin{figure}[!hbt]
  \centering
  \includegraphics[width=.8\linewidth]{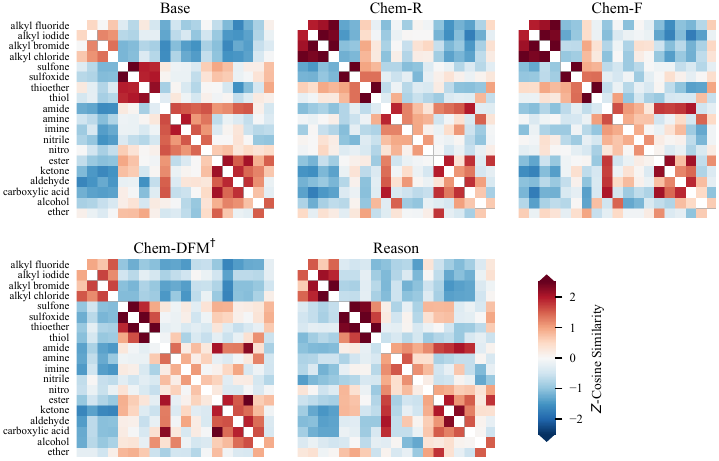}
  \caption{\textbf{Between-group $Z$-cosine similarity at the final layer.} Each cell is the cosine similarity between two functional groups' mean difference vectors, standardized within each model's panel. All panels use \base{}'s cluster order. Diagonal blank. Figure~S1 shows these panels at every sampled layer, with the same group ordering (Section~S4 of the Supporting Information). $^{\dagger}$Built on Llama-3.}
  \label{fig:clustermaps}
\end{figure}

To compare specific pairs of groups, Figure~\ref{fig:clustermaps} uses the group mean $\bar{d}_g$, the average of $d_{g,c}$ over the four chain lengths, for each of the 19 non-alkane groups. The alkane is excluded because its difference vector is zero. Because the overall level of these cosines differs between models by up to a factor of two, we report them as $Z$-cosines, standardized against the $\binom{19}{2} = 171$ between-group values of the same model and layer.  A $Z$-cosine therefore indicates how similar two groups are relative to the other group pairs in the same model, such as whether amide is closer to the carbonyl groups than to ether, but it cannot be compared across models (Supporting Information Section~S7).

\begin{table}[!hbt]
  \centering
  \caption{\textbf{Anisotropy and difference-vector similarity at layer 31.} \emph{Pairwise}: mean cosine similarity (sd) between raw embeddings of all molecule pairs. \emph{SVD}$_1$: variance share of the dominant direction of the uncentred difference vectors. \emph{Within}/\emph{between}: mean cosine of mean-centred difference vectors from the same or different groups. The uncentred values are in Table~S5. $^{\dagger}$Built on Llama-3.}
  \label{tab:geometry}
  \begin{tabular}{lcccc}
    \toprule
    & \multicolumn{2}{c}{Anisotropy} & \multicolumn{2}{c}{Mean-centered cosine} \\
    \cmidrule(lr){2-3}\cmidrule(lr){4-5}
    Model & pairwise (mean $\pm$ sd) & SVD$_1$ & within & between \\
    \midrule
    Llama-3.1-8B (Base) & $0.944 \pm 0.018$ & 0.469 & 0.754 & $-0.044$ \\
    Chem-R-8B (Chem-R) & $0.668 \pm 0.073$ & 0.349 & 0.596 & $-0.038$ \\
    Chem-R-Faithful (Chem-F) & $0.676 \pm 0.073$ & 0.353 & 0.616 & $-0.039$ \\
    ChemDFM-v1.5-8B$^{\dagger}$ (Chem-DFM) & $0.582 \pm 0.095$ & 0.378 & 0.631 & $-0.040$ \\
    R1-Distill-Llama-8B (Reason) & $0.806 \pm 0.057$ & 0.421 & 0.736 & $-0.044$ \\
    \bottomrule
  \end{tabular}
\end{table}

Across all models, functional groups cluster by chemical family, even though this grouping is not explicitly supervised during training and functional group information occupies only a small portion of the prompt template (Figure~\ref{fig:clustermaps}). The halide, sulfur, nitrogen, and carbonyl groups each form a distinct block of mutually similar groups at the final layer, with the nitrogen block weakest in \chemdfm{} and \chem{}. The halide block varies notably across models, and is tightest in \chemr{} and \chem{} and loosest in \base{}. However, because all four alkyl halides share the \emph{alkyl \ldots{}ide} naming pattern, this block could cohere partly through spelling, so the sulfur, nitrogen, and carbonyl blocks provide the clearer evidence of chemical organization. Amide, which contains both nitrogen and a carbonyl, falls in both blocks. It groups with the nitrogen family but is also more similar than average to aldehyde, ketone, ester, and carboxylic acid in all five models, as its $-$C(=O)N$-$ structure predicts.

\begin{figure}[!hbt]
  \centering
  \includegraphics[width=.8\linewidth]{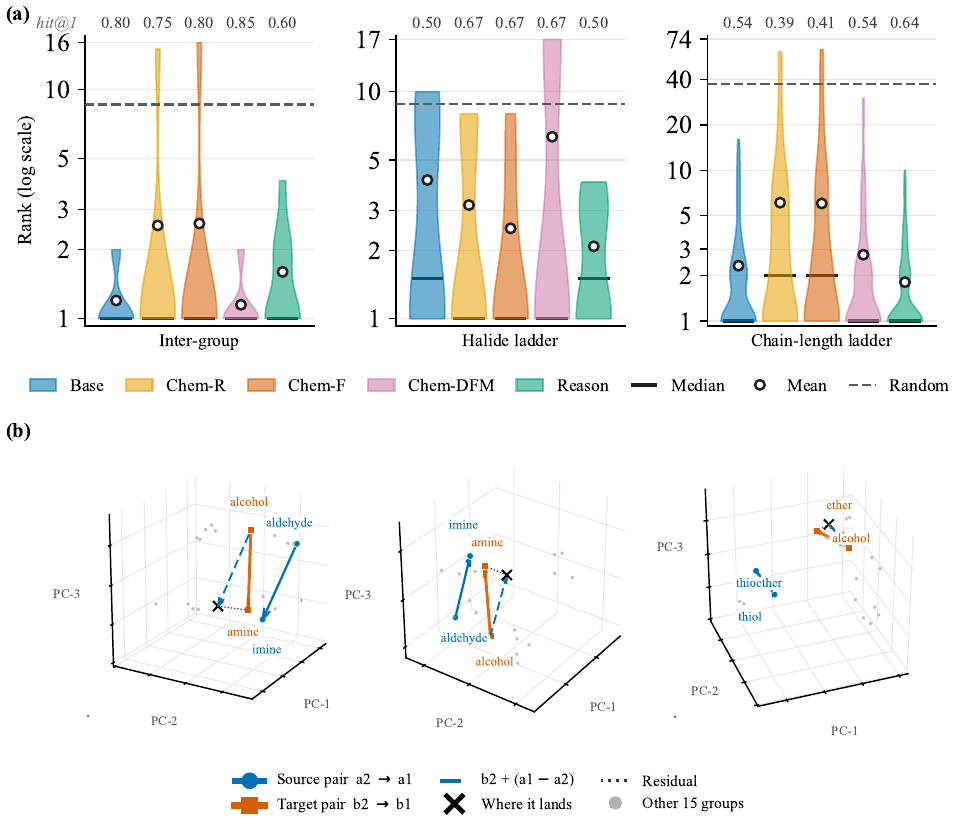}
  \caption{\textbf{Retrieval along three types of chemical difference.} (a) Rank of the correct group predicted by analogy (log scale; lower is better), with hit@1 above each model. Bars: median; circles: mean; dashed line: expected rank under random ordering. Panels use different candidate pools and are not directly comparable. (b) Three analogies for \base{} in the first three principal components.}
  \label{fig:retrieval}
\end{figure}

Chemical differences between functional groups behave as consistent offsets, so that the change from one group to another can be transferred to a third (Figure~\ref{fig:retrieval}). We test this with the analogy task for word vectors \cite{mikolov2013linguistic}. Given two pairs related by the same chemical change, $a_2 \to a_1$ and $b_2 \to b_1$, we predict $b_1$ as $b_2 + (a_1 - a_2)$. For example, thiol $-$ thioether $+$ ether should land near alcohol (Methods Section). We rank all candidates except for the three inputs by cosine similarity to the predicted vector and report hit@1, the share of trials in which $b_1$ ranks first, together with the rank of $b_1$. Figure~\ref{fig:retrieval}b illustrates this analogy for \base{}, using vectors at a single chain length projected onto the first three principal components, and Section~S5 of the Supporting Information draws the same three analogy cells for every model.

We test three types of analogy, matching the three panels of Figure~\ref{fig:retrieval}a. The \emph{inter-group} analogies  pairs functional groups related by the same substitution, such as O$\to$N or S$\to$O (20 trials). The \emph{halide ladder} relates the four alkyl halides to one another (12 trials), and the \emph{chain-length ladder} follows each group along the chain-length ladder (228 trials). The first two use the group means averaged over chain lengths, as in Figure~\ref{fig:clustermaps}, while the chain-length ladder uses the 76 individual difference vectors, one for each group and chain length. Because the candidate pools differ, each type is evaluated against its own chance rate rather than compared with the others.

Between groups, every model retrieves the correct group at median rank 1, with hit@1 from 0.600 in \reason{} to 0.850 in \chemdfm{} (Figure~\ref{fig:retrieval}a). The long upper tails for \chemr{} and \chem{} come from two trials each in the sulfur-oxidation analogy, where they rank sulfoxide 15th or 16th of 17 candidates, while \base{} and \chemdfm{} rank it second. These trials alone account for their higher mean ranks (2.55 and 2.60, against 1.20 in \base{}). Down the halide column, all five models also retrieve well above random selection probability, with hit@1 of 0.500 to 0.667. 


The chain-length ladder, with 228 trials, is the only type with enough trials to separate models, but it also tests the subtlest distinction. Adding a single CH$_2$ unit changes a molecule far less than changing its functional group, so the offsets along this axis are small and harder to recover, and hit@1 is lower than on the other two types in every model. Here, \chemr{} and \chem{}, the two chemistry-trained models built directly on \base{}, perform worst. They are the only models at median rank 2, with hit@1 of 0.395 and 0.412, against 0.544 in \base{}, 0.539 in \chemdfm{} and 0.636 in \reason{}. Their errors are off by one rank, with the top-ranked candidate being the correct functional group at an adjacent chain length. These models therefore retain the functional group but resolve chain length less precisely. We return to this trade-off in Section~\ref{sec:finetune}.

\subsection{Layer-wise Emergence of Chemical Information}
\label{sec:progress}

The previous sections showed that the last-token hidden state at the final layer preserves chemically meaningful patterns, but not where in the model those patterns arise. To examine how the linear decodability of functional groups from hidden states changes across layers, we train a logistic regression probe at each layer and analyze the results alongside the arrangement of functional groups in the embedding space. The probe classifies the last-token state into five families: hydrocarbon, halide, oxygen, nitrogen, and sulfur (Methods section). As a reference point, we use a character $n$-gram classifier, which predicts the family from short character sequences in the prompt text (such as \texttt{-ol} or \texttt{nitr}) without using any language model and reaches a balanced accuracy of 0.475. Exceeding this baseline rules out the simplest explanation that the probe reads the spelling of the name or SMILES.

\begin{figure}[!hbt]
  \centering
  \includegraphics[width=0.7\textwidth]{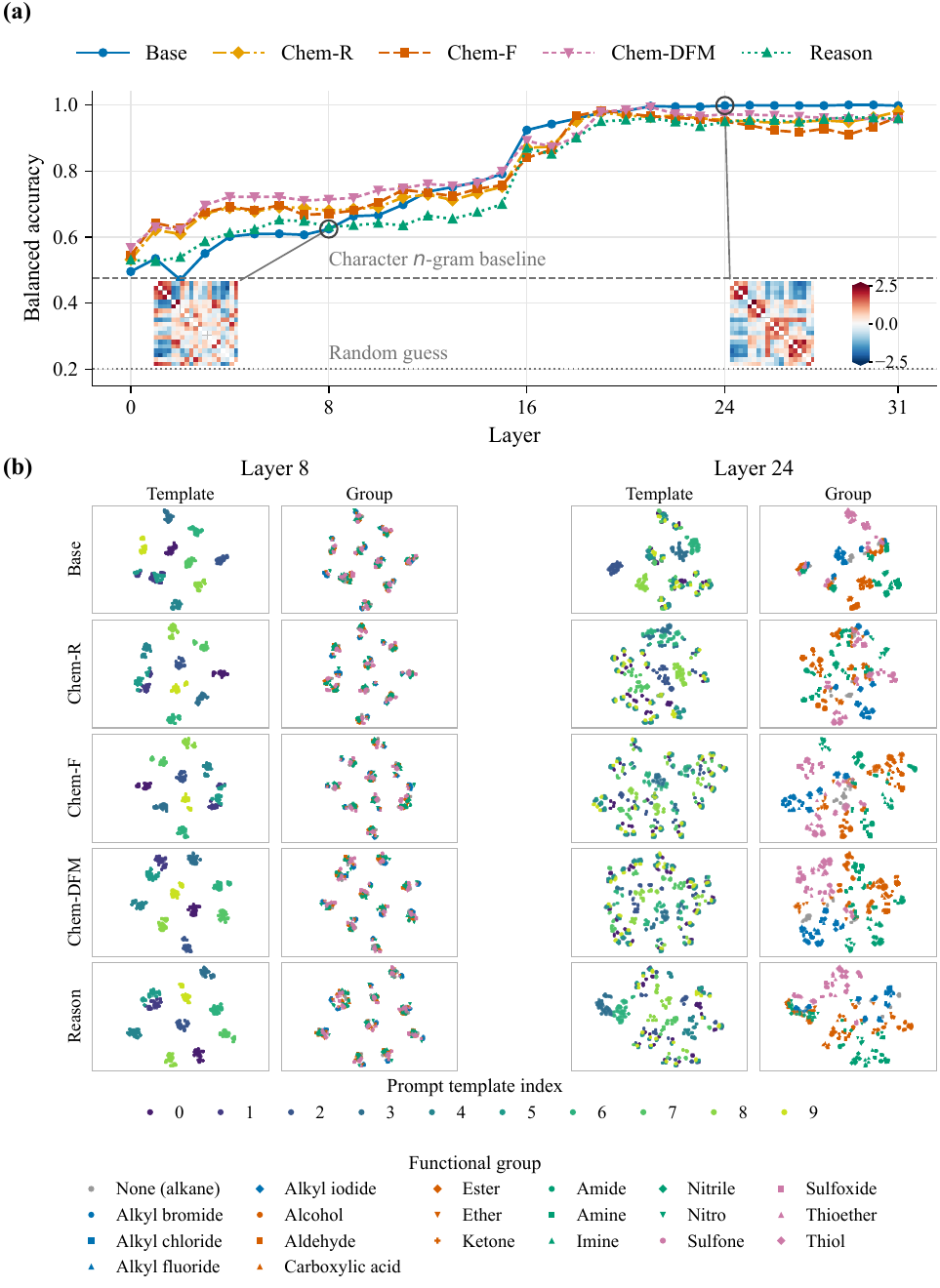}
  \caption{\textbf{Layer-wise decodability of functional-group family.} (a) Probe balanced accuracy by layer, with random guess (dotted, 0.20) and a character $n$-gram baseline (dashed, 0.475). Insets: \base{}'s between-group similarity clustermaps at layers 8 and 24, as in Figure~\ref{fig:clustermaps}. (b) $t$-SNE at layers 8 and 24, colored by template and by functional group.}
  \label{fig:depth}
\end{figure}

Throughout the first half of the network, functional-group families are only partially decodable. The balanced accuracy increases gradually from near the $n$-gram baseline in the first layers and remains below 0.80 at layer 15 in every model (Figure~\ref{fig:depth}a). Then it increases sharply, with the largest single-layer gain at layer 16 in four models and at layer 18 in \chem{}. The transition is most abrupt in \base{}, where the step from 0.791 to 0.924 at layer 16 accounts for 71\% of its rise between layers 15 and 19. By layer 19, every model reaches 0.950 to 0.983, so family information becomes nearly fully linearly decodable only after this mid-network transition. The transition occurs at a similar depth even in \chemdfm{}, which shares the architecture of the other models but not their base weights, suggesting that it is not specific to one set of pretrained weights. 


The same transition is visible without a classifier. In the $t$ -SNE projections, embeddings cluster by prompt template until layer 8, whereas after the sharp transition point at mid-layer, they cluster by functional group (layer 24), with the template-based clustering becoming less pronounced (Figure~\ref{fig:depth}b). The close-neighbor agreement, analyzed in Section~\ref{sec:finetune}, shows that this shift occurs over the same layers as the increase in probe precision (Figure~\ref{fig:crossover}). These analyzes are correlational.

The slight drop in probe accuracy for models other than \base{} may reflect how they distribute probability across competing functional-group families. The probe on \base{} is nearly deterministic, assigning almost all probability to the conventional family, whereas probes on the chemistry-tuned models assign more probability to chemically plausible alternatives. For example, at layer 31, the \base{} probe correctly classifies every amide but assigns 0.156 of its probability to the oxygen family, whereas \chemr{} assigns 0.618 to the nitrogen family and 0.325 to the oxygen family (Figure~S3). Further analysis is provided in Section~S6 of the Supporting Information.

\subsection{Fine-Tuning-Induced Representational Shifts}
\label{sec:finetune}

Although \base{} already organizes its embedding space by chemical family, fine-tuning reshapes the geometry of that space in more subtle ways. Mean pairwise cosine between raw embeddings, a measure of anisotropy, falls from 0.944 in \base{} to 0.582--0.676 in the three chemistry-trained models, and to 0.806 in \reason{} (Table~\ref{tab:geometry}). The difference vectors are also less dominated by a single shared direction. SVD$_1$ falls from 0.469 in \base{} to between 0.349 and 0.378 in the chemistry-trained models, against 0.421 in \reason{}. The mean-centered similarity within the group falls in the same order in all five models, from 0.754 in \base{} to 0.596 in \chemr{}, suggesting that it reflects this reduced anisotropy rather than the weaker functional-group identity. Without mean-centring, within-group values remain 2.1 to 2.3 times the between-group level in every model (Table~S5 in Section~S8 of the Supporting Information). Fine-tuning thus makes the representation less anisotropic, and chemistry training does so most strongly. Embeddings spread across more directions instead of crowding along a shared one, which may leave more room for chemical distinctions to be expressed as differences in direction.

\begin{figure}[!hbt]
  \centering
  \includegraphics[width=0.9\textwidth]{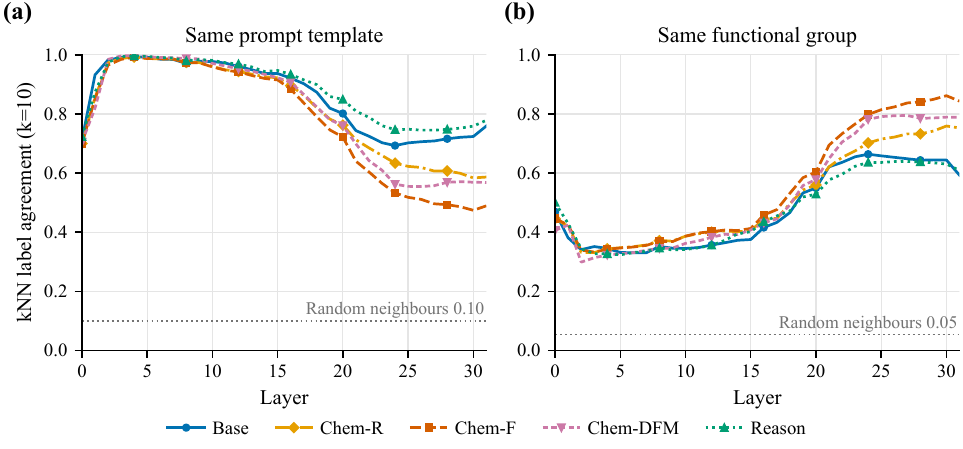}
  \caption{\textbf{Nearest-neighbour label agreement across layers.} Share of each embedding's 10 nearest neighbours by cosine similarity with (a) the same prompt template and (b) the same functional group, computed on all 920 embeddings without averaging over templates. Dotted lines: label agreement expected for random neighbours.}
  \label{fig:crossover}
\end{figure}

The same shift appears in what each embedding's nearest neighbors share. At each layer, we mean-center all 920 last-token embeddings, keeping the 10 templates of each molecule separate so that wording and chemistry can be told apart, and find each embedding's 10 nearest neighbors by cosine similarity, excluding itself. We then record the share of neighbors from the same prompt template and the share with the same functional group, averaged over all embeddings (Figure~\ref{fig:crossover}). A high template agreement means that embeddings sit closest to prompts with the same wording, whatever the molecule, while a high group agreement means that they sit closest to molecules with the same functional group, whatever the wording. Each label has its own random baseline. The probability that a randomly chosen other embedding shares the label is $\sum_c n_c(n_c-1)/n(n-1)$, where $n_c$ is the number of embeddings with the label $c$ and $n = 920$. This gives 0.099 for the template and 0.054 for the functional group.

In every model, early layers are organized by prompt template. At layer 4, template agreement is 0.993 to 0.998, while group agreement is only 0.323 to 0.345, so embeddings sit almost entirely with prompts of the same wording. In later layers, from the probe's transition onward, template agreement falls, and group agreement rises in all five models, consistent with the $t$-SNE projections, where functional groups begin to cluster together after the mid-layer point (Figure~\ref{fig:depth}b). This shift is much stronger in the chemistry-trained models. By layer 31, template agreement falls to 0.587 in \chemr{} and 0.489 in \chem{}, against 0.759 in \base{} and 0.779 in \reason{}, while group agreement rises to 0.753 and 0.843, against 0.593 and 0.609. As a result, group agreement overtakes template agreement only in the chemistry-trained models, at layer 21 in \chem{}, 22 in \chemdfm{}$^{\dagger}$ and 23 in \chemr{}, whereas in \base{} and \reason{} template agreement remains higher through the final layer. Chemistry training is thus the factor that distinguishes the models in which functional group outweighs prompt wording.

This difference parallels the behavior of the models in generation. \base{}, whose final-layer neighborhoods remain organized by wording, is also the most sensitive to wording in its answers. Its accuracy ranges from 0.196 to 0.880 across templates (Table~S4 in Section~S3 of the Supporting Information), with a single template accounting for most of its malformed responses. In \chemr{} and \chem{}, where the functional group outweighs the wording, the weakest template still reaches 0.652 and 0.685. A representation organized by functional group may thus make a model's answers less dependent on how the question is phrased. 


\section{Conclusion}
\label{sec:conclusion}

Functional-group knowledge is structured in these language models before it is spoken. Even the general-purpose Llama-3.1-8B base model, which may not be able to answer primary functional groups with high accuracy, organizes functional groups by chemical family and encodes chemically meaningful patterns in its embedding space. Its low score reflects how often it produces a usable answer, not how often it is wrong, so its chemistry is largely present but unexpressed.

Chemistry fine-tuning acts less by supplying chemical knowledge than by reorganizing it. Functional-group family becomes linearly decodable in the middle layers of every model, with or without chemistry training. What chemistry training changes is which factor dominates the representation. In the chemistry-trained models, functional group comes to outweigh the wording of the prompt, whereas in the general-purpose base and reasoning-distilled models it never does.

This distinction matters for building chemical agents, where a language model's reading of molecular structure feeds every later step, from identifying reactive sites to proposing syntheses. Output-based benchmarks cannot reveal why a model performs poorly, because they combine the chemical information encoded in its representations with its ability to express that information in the expected form. A low accuracy could mean that the relevant structural information is absent from the representation, or that it is present but not stated in a usable form. Reading the representation can distinguish these cases. For open-weight models, which agents must use whenever proprietary structures or unpublished data cannot be sent to an external API, this has a direct practical use. If the representation encodes the relevant information but the output does not, lighter interventions such as answer-format tuning, few-shot prompting, or constrained decoding may suffice in place of further chemical training. More broadly, the same analysis offers a starting point for a mechanistic understanding of how language models process chemical text, beyond what their outputs alone can show.



\section*{Acknowledgments}
The authors acknowledge the Department of Chemical and Biomolecular Engineering at the University of Nebraska--Lincoln for support.

\section*{Technology Use Disclosure}
During the preparation of this work, the authors used ChatGPT (OpenAI) and Claude (Anthropic) to assist with code development and manuscript writing. The authors reviewed, tested and edited all code and content as needed, and take full responsibility for the accuracy and integrity of the work.

\section*{Supporting information}

The Supporting Information is available free of charge and contains: Molecule Set (Section~S1); Prompt Templates (Section~S2); Generation Quality (Section~S3); Clustermap Across Layers (Section~S4); 3D Quiver Examples (Section~S5); Graded Ambiguity (Section~S6); Mean-centered Cosine and $Z$-Cosine (Section~S7); and Raw Similarity (Section~S8).

\printbibliography

\end{document}


\maketitle

\tableofcontents
\clearpage
\section{Molecule Set}
\label{app:molecule}

The probe set contains 92 small aliphatic molecules spanning 20
functional-group classes, including the unsubstituted alkane
(Table~\ref{tab:molecules}). Each class is
represented once at every chain length from three to six carbons, except
alcohol, amine, and thiol, which include both the 1- and 2-positional isomers
at each length. The alkanes serve as the reference backbone against which
each difference vector is computed.

\begin{table}[hbtp]
  \centering
  \caption{Molecule set (92 molecules). Each cell gives the IUPAC name (top)
  and SMILES (bottom) of one molecule; C, number of carbon atoms. Alcohols,
  amines and thiols include the 1- and 2-positional isomers.}
  \label{tab:molecules}
  \scriptsize
  \setlength{\tabcolsep}{4pt}
  \renewcommand{\arraystretch}{1.1}
  \newcommand{\mol}[2]{\makecell[tl]{#1\\\texttt{#2}}}
  \begin{tabular}{@{}lllll@{}}
    \toprule
    Functional group & C$_3$ & C$_4$ & C$_5$ & C$_6$ \\
    \midrule
    \multicolumn{5}{@{}l}{\emph{Hydrocarbon}} \\
    none (alkane) & \mol{Propane}{CCC} & \mol{Butane}{CCCC} & \mol{Pentane}{CCCCC} & \mol{Hexane}{CCCCCC} \\
    \addlinespace[3pt]
    \multicolumn{5}{@{}l}{\emph{Halide}} \\
    alkyl fluoride & \mol{1-Fluoropropane}{CCCF} & \mol{1-Fluorobutane}{CCCCF} & \mol{1-Fluoropentane}{CCCCCF} & \mol{1-Fluorohexane}{CCCCCCF} \\
    alkyl chloride & \mol{1-Chloropropane}{CCCCl} & \mol{1-Chlorobutane}{CCCCCl} & \mol{1-Chloropentane}{CCCCCCl} & \mol{1-Chlorohexane}{CCCCCCCl} \\
    alkyl bromide & \mol{1-Bromopropane}{CCCBr} & \mol{1-Bromobutane}{CCCCBr} & \mol{1-Bromopentane}{CCCCCBr} & \mol{1-Bromohexane}{CCCCCCBr} \\
    alkyl iodide & \mol{1-Iodopropane}{CCCI} & \mol{1-Iodobutane}{CCCCI} & \mol{1-Iodopentane}{CCCCCI} & \mol{1-Iodohexane}{CCCCCCI} \\
    \addlinespace[3pt]
    \multicolumn{5}{@{}l}{\emph{Oxygen}} \\
    alcohol (1-) & \mol{Propan-1-ol}{CCCO} & \mol{Butan-1-ol}{CCCCO} & \mol{Pentan-1-ol}{CCCCCO} & \mol{Hexan-1-ol}{CCCCCCO} \\
    alcohol (2-) & \mol{Propan-2-ol}{CC(O)C} & \mol{Butan-2-ol}{CC(O)CC} & \mol{Pentan-2-ol}{CC(O)CCC} & \mol{Hexan-2-ol}{CC(O)CCCC} \\
    ether & \mol{Methoxyethane}{CCOC} & \mol{1-Methoxypropane}{CCCOC} & \mol{1-Methoxybutane}{CCCCOC} & \mol{1-Methoxypentane}{CCCCCOC} \\
    aldehyde & \mol{Propanal}{CCC=O} & \mol{Butanal}{CCCC=O} & \mol{Pentanal}{CCCCC=O} & \mol{Hexanal}{CCCCCC=O} \\
    ketone & \mol{Propan-2-one}{CC(=O)C} & \mol{Butan-2-one}{CC(=O)CC} & \mol{Pentan-2-one}{CC(=O)CCC} & \mol{Hexan-2-one}{CC(=O)CCCC} \\
    carboxylic acid & \mol{Propanoic acid}{CCC(O)=O} & \mol{Butanoic acid}{CCCC(O)=O} & \mol{Pentanoic acid}{CCCCC(O)=O} & \mol{Hexanoic acid}{CCCCCC(O)=O} \\
    ester & \mol{Methyl ethanoate}{CC(=O)OC} & \mol{Methyl propanoate}{CCC(=O)OC} & \mol{Methyl butanoate}{CCCC(=O)OC} & \mol{Methyl pentanoate}{CCCCC(=O)OC} \\
    \addlinespace[3pt]
    \multicolumn{5}{@{}l}{\emph{Nitrogen}} \\
    amine (1-) & \mol{Propan-1-amine}{CCCN} & \mol{Butan-1-amine}{CCCCN} & \mol{Pentan-1-amine}{CCCCCN} & \mol{Hexan-1-amine}{CCCCCCN} \\
    amine (2-) & \mol{Propan-2-amine}{CC(N)C} & \mol{Butan-2-amine}{CC(N)CC} & \mol{Pentan-2-amine}{CC(N)CCC} & \mol{Hexan-2-amine}{CC(N)CCCC} \\
    imine & \mol{Propan-1-imine}{CCC=N} & \mol{Butan-1-imine}{CCCC=N} & \mol{Pentan-1-imine}{CCCCC=N} & \mol{Hexan-1-imine}{CCCCCC=N} \\
    nitrile & \mol{Propanenitrile}{CCC\#N} & \mol{Butanenitrile}{CCCC\#N} & \mol{Pentanenitrile}{CCCCC\#N} & \mol{Hexanenitrile}{CCCCCC\#N} \\
    amide & \mol{Propanamide}{CCC(N)=O} & \mol{Butanamide}{CCCC(N)=O} & \mol{Pentanamide}{CCCCC(N)=O} & \mol{Hexanamide}{CCCCCC(N)=O} \\
    nitro & \mol{1-Nitropropane}{CCC[N+](=O)[O-]} & \mol{1-Nitrobutane}{CCCC[N+](=O)[O-]} & \mol{1-Nitropentane}{CCCCC[N+](=O)[O-]} & \mol{1-Nitrohexane}{CCCCCC[N+](=O)[O-]} \\
    \addlinespace[3pt]
    \multicolumn{5}{@{}l}{\emph{Sulfur}} \\
    thiol (1-) & \mol{Propane-1-thiol}{CCCS} & \mol{Butane-1-thiol}{CCCCS} & \mol{Pentane-1-thiol}{CCCCCS} & \mol{Hexane-1-thiol}{CCCCCCS} \\
    thiol (2-) & \mol{Propane-2-thiol}{CC(S)C} & \mol{Butane-2-thiol}{CC(S)CC} & \mol{Pentane-2-thiol}{CC(S)CCC} & \mol{Hexane-2-thiol}{CC(S)CCCC} \\
    thioether & \mol{(Methylsulfanyl)ethane}{CCSC} & \mol{1-(Methylsulfanyl)propane}{CCCSC} & \mol{1-(Methylsulfanyl)butane}{CCCCSC} & \mol{1-(Methylsulfanyl)pentane}{CCCCCSC} \\
    sulfoxide & \mol{Methyl ethyl sulfoxide}{CS(=O)CC} & \mol{Methyl propyl sulfoxide}{CS(=O)CCC} & \mol{Methyl butyl sulfoxide}{CS(=O)CCCC} & \mol{Methyl pentyl sulfoxide}{CS(=O)CCCCC} \\
    sulfone & \mol{Methyl ethyl sulfone}{CS(=O)(=O)CC} & \mol{Methyl propyl sulfone}{CS(=O)(=O)CCC} & \mol{Methyl butyl sulfone}{CS(=O)(=O)CCCC} & \mol{Methyl pentyl sulfone}{CS(=O)(=O)CCCCC} \\
    \bottomrule
  \end{tabular}
\end{table}

\newpage
\section{Prompt Templates}
\label{app:prompt}

Each molecule's IUPAC name and SMILES string are inserted into the ten
prompt templates listed in Table~\ref{tab:templates}. Every template ends mid-sentence, so that the natural continuation is the name of the molecule's primary functional group.

\begin{table}[htbp]
  \centering
  \caption{Prompt templates. Each is filled with the IUPAC name and SMILES
  of all 92 molecules, giving 920 prompts per model. Templates differ in
  word order, molecule position and target noun (\emph{group}, \emph{class},
  \emph{identity}) and end mid-sentence so that the functional-group name is
  the natural continuation.}
  \label{tab:templates}
  \footnotesize
  \setlength{\tabcolsep}{4pt}
  \begin{tabular}{@{}rp{0.90\linewidth}@{}}
    \toprule
    & Template \\
    \midrule
    1 & \ttfamily \{iupac\_name\} (\{smiles\}) contains the functional group \\
    2 & \ttfamily The primary functional group in \{iupac\_name\} (\{smiles\}) is \\
    3 & \ttfamily For \{iupac\_name\} (\{smiles\}), the functional-group class is \\
    4 & \ttfamily \{iupac\_name\} (\{smiles\}) is classified as a molecule containing a \\
    5 & \ttfamily The characteristic functional group of \{iupac\_name\} (\{smiles\}) is \\
    6 & \ttfamily The main functional group present in \{iupac\_name\} (\{smiles\}) is \\
    7 & \ttfamily When classified by functional group, \{iupac\_name\} (\{smiles\}) is a \\
    8 & \ttfamily \{iupac\_name\} (\{smiles\}) can be categorized by the functional group \\
    9 & \ttfamily The functional-group identity of \{iupac\_name\} (\{smiles\}) is \\
    10 & \ttfamily For the molecule \{iupac\_name\} (\{smiles\}), the defining functional group is \\
    \bottomrule
  \end{tabular}
\end{table}

\newpage
\section{Generation Quality}
\label{app:generation}

Each greedy continuation is assigned to exactly one of five outcomes:

\begin{itemize}
  \item \textbf{Correct}: the answer names exactly one functional group, and it matches the target.
  \item \textbf{Wrong}: the answer names exactly one functional group, and it does not match.
  \item \textbf{Malformed}: the continuation does not take the form of an answer. It is an \emph{option list} (two or more enumeration markers), a \emph{blank-fill} (three or more underscores), a \emph{preamble} (it opens with reasoning such as \emph{Okay, so}), a \emph{hedge} (it names several groups as alternatives, joined by \emph{or}, \emph{and} or a slash), or a \emph{self-retraction} (e.g.\ \emph{but wait}).
  \item \textbf{Underspecified}: the answer names no specific group, only a superclass that is true of four or five groups at once (e.g.\ \emph{carbonyl}, \emph{halide}, \emph{halogen}, \emph{haloalkane}).
  \item \textbf{Non-answer}: the answer names neither a functional group nor a superclass.
\end{itemize}

Strict accuracy is the number of correct answers divided by the
$N = 920$ prompts per model, and lenient accuracy adds the underspecified answers to the numerator:
\begin{equation}
  \mathrm{Acc}_{\text{strict}} = \frac{n_{\text{correct}}}{N},
  \qquad
  \mathrm{Acc}_{\text{lenient}} = \frac{n_{\text{correct}} + n_{\text{underspecified}}}{N}.
  \label{eq:freegen}
\end{equation}

Both use all prompts as the denominator, so a model is not rewarded for
declining to answer. Underspecified answers are credited without checking that the superclass contains the target functional group. 



Table~\ref{tab:noncommitment} breaks down the types on which each model
names no functional-group class. The total non-commitment rate varies
widely, from 0.081 for \chemr{} to 0.364 for \base{}, and the dominant type differs by model. \base{} and \chemdfm{} decline to commit answers at similar rates (0.364 and 0.335) but by different routes: \base{} spreads across all three types, with the largest share of malformed responses of any model (0.132), whereas \chemdfm{} places 0.85 of its non-commitments in non-answers, which describe the molecule or its reactivity without naming a group. Most of \base{}'s malformed responses come from a single template, with template~9 accounting for 0.57 of them. \chem{} is the only model whose non-commitments are mostly underspecified (0.093 of 0.142), typically naming the carbonyl group without resolving it to a specific class. Chemistry fine-tuning thus nearly eliminates format failures (0.002 to 0.007 against 0.132 for \base{}) but does not by itself ensure that the model commits to a single class.

\begin{table}[!hbtp]
  \centering
  \caption{Response outcomes by type. \emph{Answered} prompts name exactly
  one class (correct or wrong); the remaining three types are
  non-commitments, with one example of each. Rate is the share of all 920
  prompts, so the four rates sum to one for each model. Each example is the
  first instance by template index and molecule name, shown verbatim with
  whitespace collapsed. $^{\dagger}$Built on Llama-3 rather than
  Llama-3.1.}
  \label{tab:noncommitment}
  \small
  \begin{tabular}{@{}lcp{0.56\linewidth}@{}}
    \toprule
    Model & Rate & First instance, verbatim \\
    \midrule
    \multicolumn{3}{l}{\emph{Malformed} --- an option list, a blank-fill, a preamble, a hedge or a self-retraction} \\
    \quad Llama-3.1-8B (Base) & 0.132 & \texttt{(s) of: A. Alkane} \\
    \quad Chem-R-8B (Chem-R) & 0.007 & \texttt{bromine atom and a hydrocarbon chain. The} \\
    \quad Chem-R-Faithful (Chem-F) & 0.007 & \texttt{bromine atom and a hydrocarbon chain. The} \\
    \quad ChemDFM-v1.5-8B$^{\dagger}$ (Chem-DFM) & 0.002 & \texttt{found in both alkanes and sulfones.} \\
    \quad R1-Distill-Llama-8B (Reason) & 0.015 & \texttt{(s) of a thiol, but wait,} \\
    \midrule
    \multicolumn{3}{l}{\emph{Underspecified} --- only a superclass, true of four or five classes at once} \\
    \quad Llama-3.1-8B (Base) & 0.083 & \texttt{of an alkyl halide. The alkyl} \\
    \quad Chem-R-8B (Chem-R) & 0.013 & \texttt{carbonyl group (C=O) and} \\
    \quad Chem-R-Faithful (Chem-F) & 0.093 & \texttt{of a carbonyl group (C=O} \\
    \quad ChemDFM-v1.5-8B$^{\dagger}$ (Chem-DFM) & 0.047 & \texttt{(carbonyl group).} \\
    \quad R1-Distill-Llama-8B (Reason) & 0.058 & \texttt{of a substituted alkyl halide. The structure} \\
    \midrule
    \multicolumn{3}{l}{\emph{Non-answer} --- no group named: describes the molecule, restates the prompt} \\
    \quad Llama-3.1-8B (Base) & 0.149 & \texttt{of a fluorocarbon. It is a color} \\
    \quad Chem-R-8B (Chem-R) & 0.061 & \texttt{\textquotedbl{}Sulfur\textquotedbl{} with the SMILES pattern} \\
    \quad Chem-R-Faithful (Chem-F) & 0.042 & \texttt{functional group that can be involved in various chemical reactions} \\
    \quad ChemDFM-v1.5-8B$^{\dagger}$ (Chem-DFM) & 0.286 & \texttt{that is the target for hydrolysis by the} \\
    \quad R1-Distill-Llama-8B (Reason) & 0.099 & \texttt{-SMe, which is a good leaving group} \\
    \bottomrule
  \end{tabular}
\end{table}

This classification requires deciding whether a free-form continuation
names one of the target classes. Each continuation is therefore parsed by a fixed sequence of rules, and the order in which they are applied affects the outcome:

\begin{enumerate}
  \item The molecule's IUPAC name is removed from the continuation, since
    names such as \emph{Pentanal} or \emph{1-Bromopentane} encode the
    answer and an echo of the prompt would otherwise score as correct.
  \item The format checks for \emph{malformed} run next, before any class
    name is matched. An option list that happens to include the gold class
    is therefore malformed, not correct.
  \item The remaining text is truncated to its first clause (up to the
    first period or newline), so the answer is read where the model first
    commits to one.
  \item The clause is matched against a lexicon of accepted names and
    synonyms for the 20 classes, including element names and formula forms
    (so \emph{the bromine atom} names an alkyl bromide and \texttt{-SH} a
    thiol). Longer names take precedence over shorter ones
    (\emph{alkyl bromide} over \emph{bromide}), and matches require word
    boundaries (so \emph{bromo} does not match inside \emph{Bromopentane}).
    If nothing matches, the clause is matched once more with CamelCase
    split into words, since \chemr{} writes names such as
    \texttt{CarboxylicAcid}.
  \item If the clause names more than one functional group, the text between mentions decides. A coordinator (\emph{or}, \emph{and}, a slash) or a gap of more than two words marks a list of alternatives, which is malformed. Adjacent names are read as a compound resolved to the last one, so \emph{carboxylic acid ester} is an ester.
\end{enumerate}

Accuracy is not evenly distributed over the ten templates. Table~\ref{tab:template_accuracy} gives strict accuracy per template for each model, with the mean, standard deviation and range beneath. \base{} spans $0.196$ to $0.880$, a range of $0.685$, whereas no template drops \chemr{} below $0.652$ or \chem{} below $0.685$. No single template is weakest in every model: template~9 carries the minimum for \base{} and \chemr{}, template~4 for \chem{} and \reason{}, and template~8 for \chemdfm{}, so the spread does not reduce to one phrasing that all five models find hard.

\begin{table}[!hbt]
  \centering
  \caption{Strict free-generation accuracy by prompt template. Each cell
  covers 92 prompts, one per molecule. Templates are numbered as in
  Table~\ref{tab:templates}. $^{\dagger}$Built on Llama-3 rather than
  Llama-3.1.}
  \label{tab:template_accuracy}
  \begin{tabular}{lccccc}
    \toprule
    & \multicolumn{5}{c}{Strict accuracy, 92 prompts per cell} \\
    \cmidrule(lr){2-6}
    Template & Base & Chem-R & Chem-F & Chem-DFM$^{\dagger}$ & Reason \\
    \midrule
    1 & 0.630 & 0.978 & 0.967 & 0.554 & 0.707 \\
    2 & 0.609 & 1.000 & 0.870 & 0.717 & 0.837 \\
    3 & 0.674 & 0.957 & 1.000 & 0.707 & 0.815 \\
    4 & 0.565 & 0.696 & 0.685 & 0.641 & 0.587 \\
    5 & 0.489 & 0.978 & 0.826 & 0.652 & 0.837 \\
    6 & 0.543 & 1.000 & 0.880 & 0.750 & 0.902 \\
    7 & 0.630 & 0.793 & 0.717 & 0.587 & 0.717 \\
    8 & 0.630 & 0.967 & 0.935 & 0.337 & 0.761 \\
    9 & 0.196 & 0.652 & 0.717 & 0.652 & 0.804 \\
    10 & 0.880 & 1.000 & 0.837 & 0.750 & 0.924 \\
    \midrule
    \emph{Mean} & 0.585 & 0.902 & 0.843 & 0.635 & 0.789 \\
    \emph{SD} & 0.172 & 0.135 & 0.109 & 0.123 & 0.100 \\
    \emph{Min} & 0.196 & 0.652 & 0.685 & 0.337 & 0.587 \\
    \emph{Max} & 0.880 & 1.000 & 1.000 & 0.750 & 0.924 \\
    \bottomrule
  \end{tabular}
\end{table}


\newpage
\section{Clustermap Across Layers}
\label{app:clustermap}

\begin{figure}[!hbtp]
  \centering
  \includegraphics[width=0.99\linewidth]{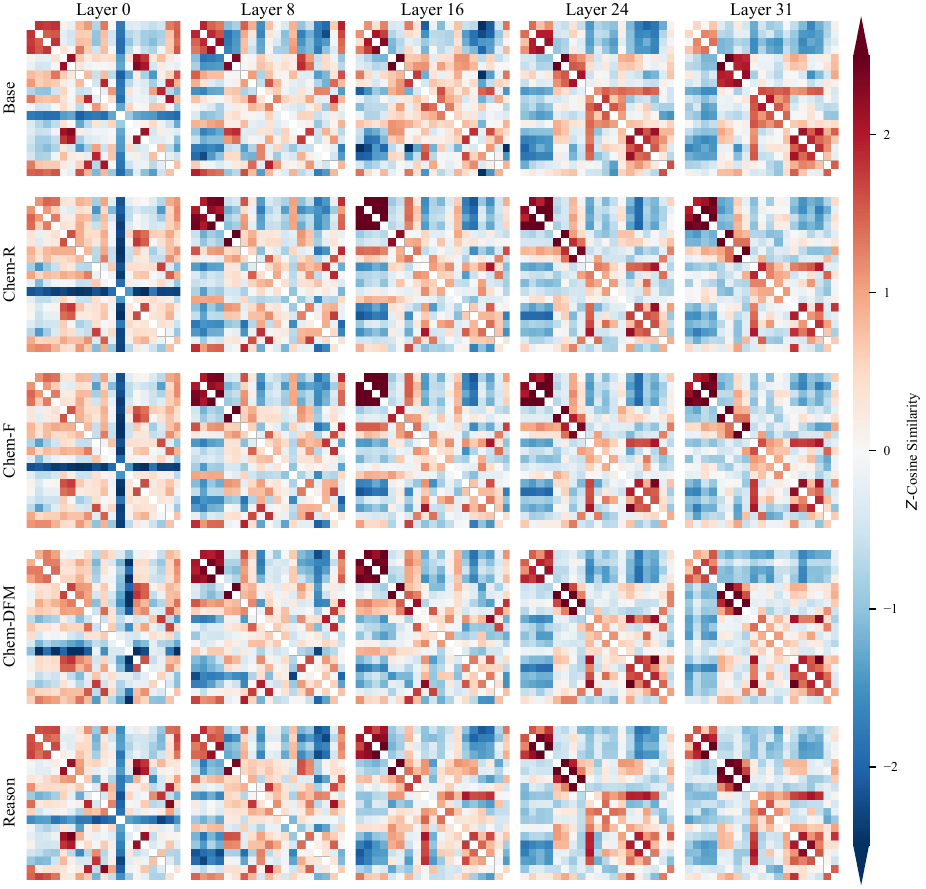}
  \caption{Between-group clustermaps for each checkpoint (rows) at layers 0, 8, 16, 24, and 31 (columns). Each cell is the cosine similarity between two functional groups' mean difference vectors, standardized within each panel ($Z$-cosine, Equation~\ref{eq:zcos}) and drawn on a shared color scale. The alkane reference is excluded. Leaf order is the same as in the layer-31 clustermaps of the main text.}
  \label{fig:clustergrid}
\end{figure}

Figure~\ref{fig:clustergrid} shows how the between-group similarity
structure in Figure~2 of the main text develops with depth. All panels use the same leaf order as that figure, so rows and columns can be compared directly across layers and models. At layer~0, only the halide and sulfur families form clusters; the other
groups remain intermixed. Clusters for the remaining heteroatom families emerge in the middle layers and are sharp by layer~24, after which the structure changes little up to the final layer.

\newpage
\section{3D Quiver Examples}
\label{app:3d_quiver}

\begin{figure}[!hbt]
  \centering
  \includegraphics[width=0.90\linewidth]{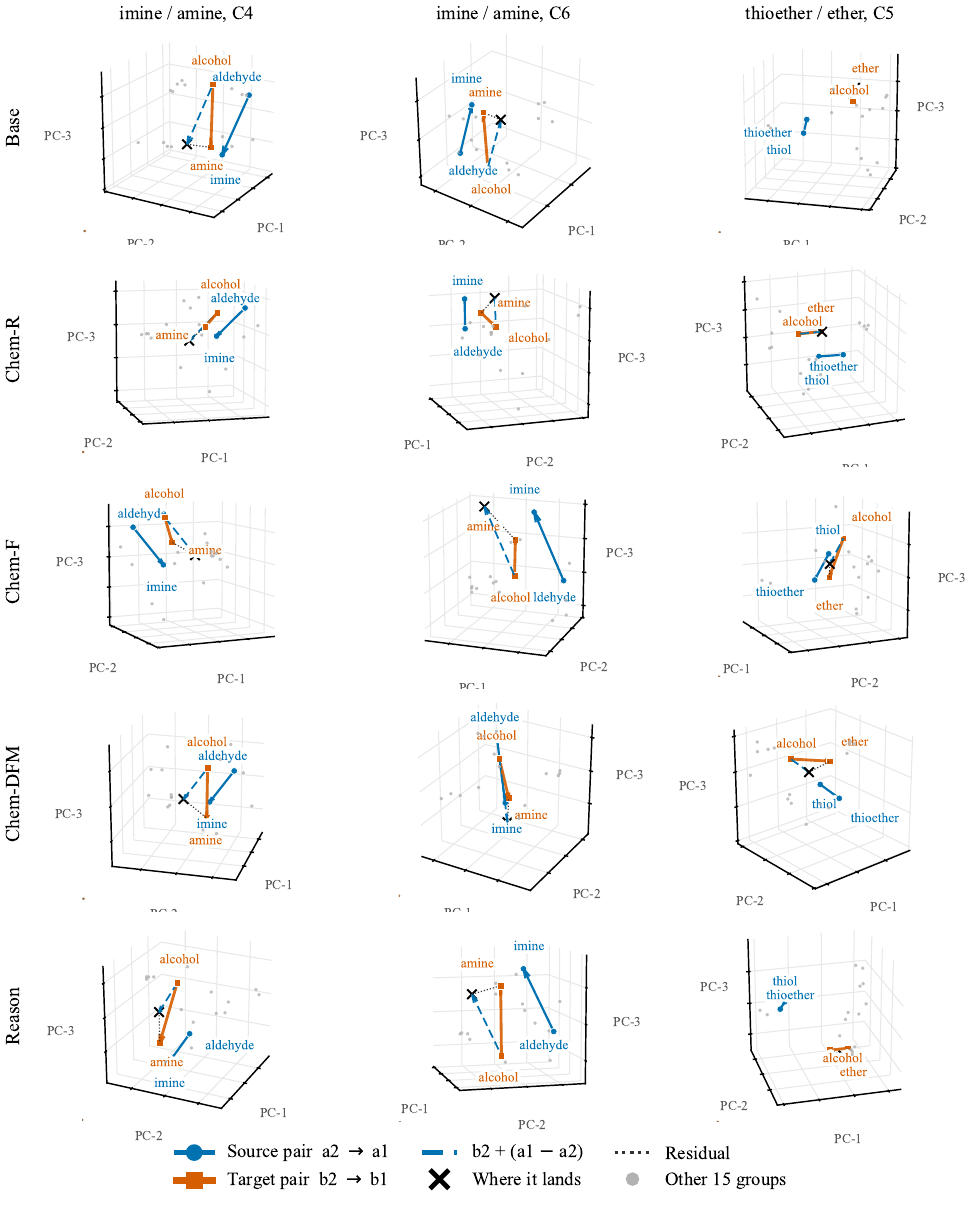}
  \caption{The three analogy examples of Figure~3b across models.}
  \label{fig:quivergrid}
\end{figure}

Figure~\ref{fig:quivergrid} extends the three analogy examples of
Figure~3b in the main text to every checkpoint. Each panel uses its own projection, so only the placement of points within a panel is comparable. 

\newpage
\section{Graded Ambiguity}
\label{app:ambiguity}

\begin{figure}[!hbtp]
  \centering
  \includegraphics[width=0.80\linewidth]{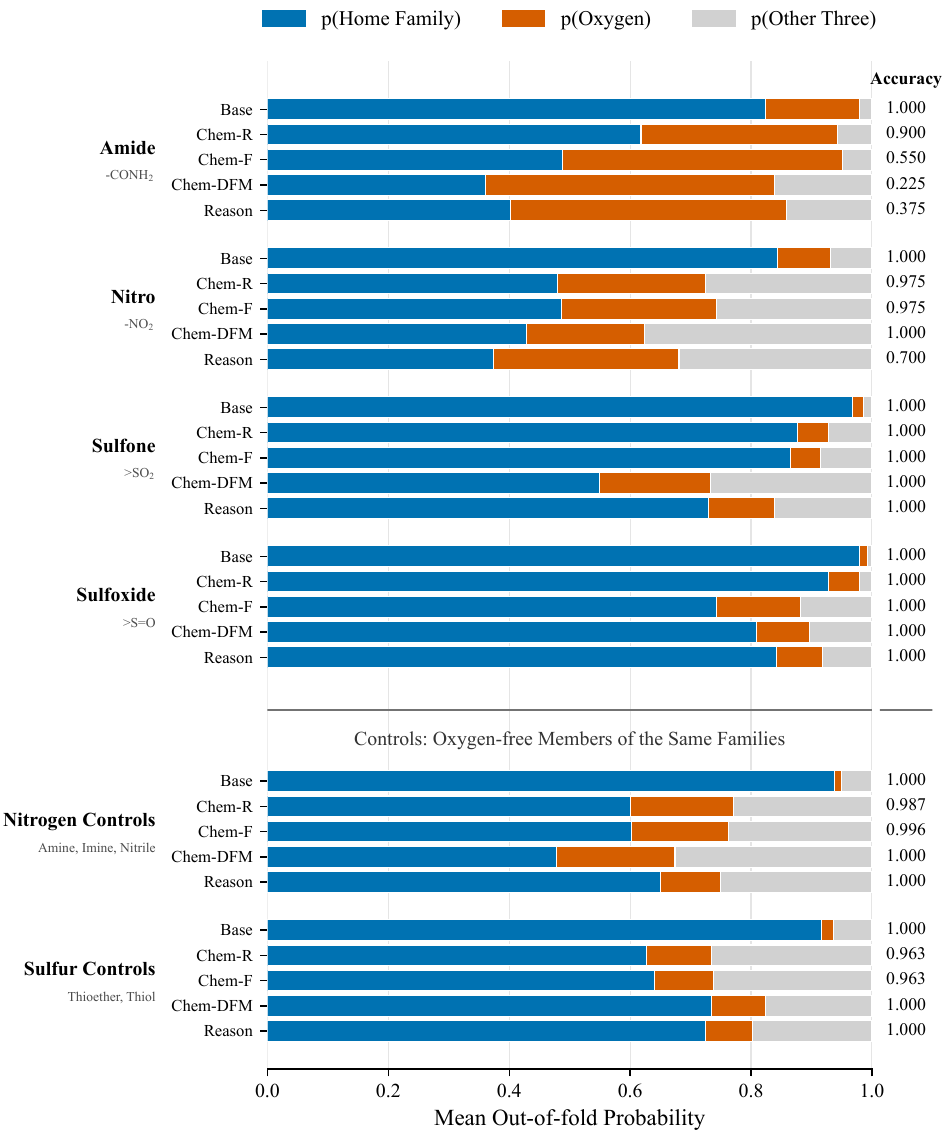}
  \caption{Family probabilities from the layer-31 probe for the four convention-labelled groups and their oxygen-free controls. Each bar shows the probe's out-of-fold probabilities under leave-one-group-out cross-validation, split into the home family, oxygen, and the remaining families; the segments of each bar sum to 1. Each control row averages its member groups. The right column gives probe accuracy of each group.}
  \label{fig:mass}
\end{figure}

Amide, nitro, sulfoxide, and sulfone contain oxygen but are filed under nitrogen or sulfur by convention, so a probe could reasonably divide its prediction between the two families. The family probe outputs a probability for each heteroatom family, and Figure~\ref{fig:mass} shows how the layer-31 probe divides this probability among the families for these four groups, alongside oxygen-free controls from the same families.
In \base{}, the probe classifies all four groups correctly (accuracy $1.000$), yet for amide it assigns a probability of $0.156$ to oxygen, compared with at most $0.018$ for any of the three nitrogen controls. Accuracy, which counts only the most probable family, scores this prediction as simply correct and does not reflect the split.

The family probabilities are directly comparable among \base{}, \chemr{} and \chem{}, but not between these and \chemdfm{} or \reason{}. The probe's regularization strength is selected separately for each model by group-aware inner cross-validation, and a stronger penalty yields flatter probability distributions. At layer~31, \base{}, \chemr{} and \chem{} all select an inverse penalty of $C = 0.001$ (in 18, 19 and 16 of the 19 folds, respectively), whereas \chemdfm{} and \reason{} select $C = 0.0001$ (in all 19 folds). The plotted values are the raw out-of-fold probabilities without temperature correction, so differences between the two sets of models partly reflect this choice of penalty. Accuracy depends only on the most probable family and is unaffected.

The probe on \base{} is close to deterministic: it assigns nearly all of its probability to the conventional family, even for groups that contain a second heteroatom. The probes on the chemistry-tuned models spread their probability more evenly between the two defensible families. Among the three models that share the same regularization ($C = 0.001$), the probability assigned to oxygen for sulfoxide rises from $0.014$ in \base{} to $0.051$ in \chemr{} and $0.140$ in \chem{}. \chemdfm{} and \reason{} show flatter distributions still, but their probes use a stronger penalty ($C = 0.0001$), which by itself flattens the probabilities, so their values cannot be separated from that choice. Where accuracy falls below ceiling, as for amide in \chemdfm{} ($0.225$), the lost probability goes mainly to oxygen, the chemically defensible alternative, rather than to unrelated families. A lower accuracy on these groups therefore reflects a representation that encodes both of their heteroatoms, not a probe that has failed.

\newpage
\section{Mean-centered Cosine and \texorpdfstring{$Z$}{Z}-Cosine}
\label{app:cosines}

This section defines the two similarity measures used in the main text: the mean-centered cosine reported in Table~2, and the $Z$-cosine plotted in Figure~2. Both start from the 76 difference vectors $d_{g,c}$ defined in Equation~1 of the main text, one for each of the 19 non-alkane functional groups $g$ at each of the four carbon counts $c \in \{3, 4, 5, 6\}$. We write $\mathcal{D} = \{d_{g,c}\}$ for this set and $\cos(u,v) = \langle u, v\rangle / (\lVert u\rVert \lVert v\rVert)$ for cosine similarity.

Both measures correct for the same problem. Every difference vector carries a large component shared by all functional groups, so uncentered cosines between different groups are high (0.32 to 0.40 on average at layer 31) regardless of training strategy. The two measures remove this shared component in different ways. The mean-centered cosine removes it from the vectors before the cosine is taken, whereas the $Z$-cosine removes it from the similarities afterward.

\paragraph{Mean-centered cosine.}
We subtract the average of all 76 difference vectors from each one,
\begin{equation}
  \bar{d}_{\mathrm{all}} = \frac{1}{|\mathcal{D}|} \sum_{d \in \mathcal{D}} d,
  \qquad
  \tilde{d} = d - \bar{d}_{\mathrm{all}},
  \label{eq:center}
\end{equation}
and then average the cosines of the centered vectors over same-group and cross-group pairs,
\begin{equation}
  s_{\mathrm{within}} = \frac{1}{|P_{\mathrm{W}}|} \sum_{(i,j) \in P_{\mathrm{W}}}
    \cos\bigl(\tilde{d}_i, \tilde{d}_j\bigr),
  \qquad
  s_{\mathrm{between}} = \frac{1}{|P_{\mathrm{B}}|} \sum_{(i,j) \in P_{\mathrm{B}}}
    \cos\bigl(\tilde{d}_i, \tilde{d}_j\bigr).
  \label{eq:withinbetween}
\end{equation}
Here $P_{\mathrm{W}}$ contains the pairs from the same functional group at different carbon counts, giving $|P_{\mathrm{W}}| = 19 \times \binom{4}{2} = 114$, and $P_{\mathrm{B}}$ contains the pairs from different functional groups, giving the remaining $|P_{\mathrm{B}}| = \binom{76}{2} - 114 = 2736$. These are the \emph{within} and \emph{between} columns of Table~2. Centering brings the between-group mean to $-0.044$ to $-0.038$ at layer 31, while the within-group mean remains between 0.596 and 0.754.

\paragraph{$Z$-cosine.}
We first average each group's difference vectors over carbon counts, giving the group mean $\bar{d}_g$ of the main text,
\begin{equation}
  \bar{d}_g = \frac{1}{4} \sum_{c=3}^{6} d_{g,c},
  \qquad
  S_{gh} = \cos\bigl(\bar{d}_g, \bar{d}_h\bigr),
  \label{eq:groupcos}
\end{equation}
which gives one similarity $S_{gh}$ for each of the $\binom{19}{2} = 171$ pairs of groups. We then standardize these similarities by their mean and standard deviation within each model and layer,
\begin{equation}
  Z_{gh} = \frac{S_{gh} - \operatorname{mean}(S_{\mathrm{off}})}
                {\operatorname{sd}(S_{\mathrm{off}})},
  \qquad
  S_{\mathrm{off}} = \{ S_{gh} : g < h \}.
  \label{eq:zcos}
\end{equation}
A $Z$-cosine is therefore expressed in standard deviations of that model's own between-group similarities rather than in cosine units. A positive value means that two groups are more similar than the average pair of groups in the same model, and a negative value means they are less similar. Because each panel of Figure~2 is standardized separately, the panels can share one colour scale, but values should be compared only within a panel.

\newpage
\section{Raw Similarity}
\label{app:geometry_raw}

Table~\ref{tab:geometry_raw} gives the uncentered counterpart of the within- and between-group columns of Table~2 in the main text, computed in the same way but without subtracting the mean difference vector (Equation~\ref{eq:center}). Without centering, similarities are high throughout, especially within groups ($0.74$ to $0.85$). Chemistry fine-tuning lowers both levels somewhat: within-group similarity falls from $0.854$ in \base{} to $0.735$ to $0.768$ in the chemistry-tuned models, and between-group similarity from $0.399$ to $0.324$ to $0.345$. Because these high levels are dominated by a component shared by all difference vectors (Section~\ref{app:cosines}), they compress the differences between groups and between models. Table~2 therefore reports mean-centered values, which remove this shared component and make the comparison easier to read.


\begin{table}[!hbt]
  \centering
  \caption{Within-group, between-group and null cosine similarity of
  difference vectors at layer 31, without mean-centering. Cosines are taken
  over individual difference vectors (one per group and chain length).
  \emph{Within} averages over pairs from the same group, \emph{between}
  over pairs from different groups, and \emph{null} over randomly sampled
  cross-group pairs, so it matches \emph{between} by construction.
  $^{\dagger}$Built on Llama-3 rather than Llama-3.1.}
  \label{tab:geometry_raw}
  \begin{tabular}{lccc}
    \toprule
    & \multicolumn{3}{c}{Uncentered cosine, over individual diff vectors} \\
    \cmidrule(lr){2-4}
    Model & within & between & null (mean $\pm$ sd) \\
    \midrule
    Llama-3.1-8B (Base) & 0.854 & 0.399 & $0.399 \pm 0.166$ \\
    Chem-R-8B (Chem-R) & 0.735 & 0.324 & $0.325 \pm 0.112$ \\
    Chem-R-Faithful (Chem-F) & 0.750 & 0.327 & $0.328 \pm 0.111$ \\
    ChemDFM-v1.5-8B$^{\dagger}$ (Chem-DFM) & 0.768 & 0.345 & $0.345 \pm 0.112$ \\
    R1-Distill-Llama-8B (Reason) & 0.832 & 0.368 & $0.369 \pm 0.138$ \\
    \bottomrule
  \end{tabular}
\end{table}